\documentclass[conference]{IEEEtran}
\IEEEoverridecommandlockouts

\usepackage{cite}
\usepackage{amsmath,amssymb,amsfonts}
\usepackage{graphicx}
\usepackage{textcomp}
\usepackage{xcolor}
\usepackage{svg}
\usepackage{algorithm}
\usepackage{algpseudocode}
\usepackage{booktabs}
\usepackage[hidelinks]{hyperref}

\def\BibTeX{{\rm B\kern-.05em{\sc i\kern-.025em b}\kern-.08em
    T\kern-.1667em\lower.7ex\hbox{E}\kern-.125emX}}

\begin{document}

\title{Push-Placement: A Hybrid Approach Integrating Prehensile and Non-Prehensile Manipulation for Object Rearrangement}

\author{
\IEEEauthorblockN{Majid Sadeghinejad\textsuperscript{*}, Arman Barghi\textsuperscript{*}, Hamed Hosseini, Mehdi Tale Masouleh, Ahmad Kalhor}
\IEEEauthorblockA{
\textit{Human and Robot Interaction Laboratory}\\
\textit{School of Electrical and Computer Engineering}\\
\textit{University of Tehran}\\
Tehran, Iran\\
Emails: \{majidsadeghinejad, arman.barghi, hosseini.hamed, m.t.masouleh, akalhor\}@ut.ac.ir}
\thanks{\textsuperscript{*}Equal contribution}
}

\maketitle

\begin{abstract}
Efficient tabletop rearrangement remains challenging due to collisions and the need for temporary buffering when target poses are obstructed. Prehensile pick-and-place provides precise control but often requires extra moves, whereas non-prehensile pushing can be more efficient but suffers from complex, imprecise dynamics. This paper proposes \emph{push-placement}, a hybrid action primitive that uses the grasped object to displace obstructing items while being placed, thereby reducing explicit buffering. The method is integrated into a physics-in-the-loop Monte Carlo Tree Search (MCTS) planner and evaluated in the PyBullet simulator. Empirical results show push-placement reduces the manipulator travel cost by up to \textbf{11.12\%} versus a baseline MCTS planner and \textbf{8.56\%} versus dynamic stacking. These findings indicate that hybrid prehensile/non-prehensile action primitives can substantially improve efficiency in long-horizon rearrangement tasks.
\end{abstract}


\begin{figure}[t]
    \centering
    \includegraphics[width=\linewidth]{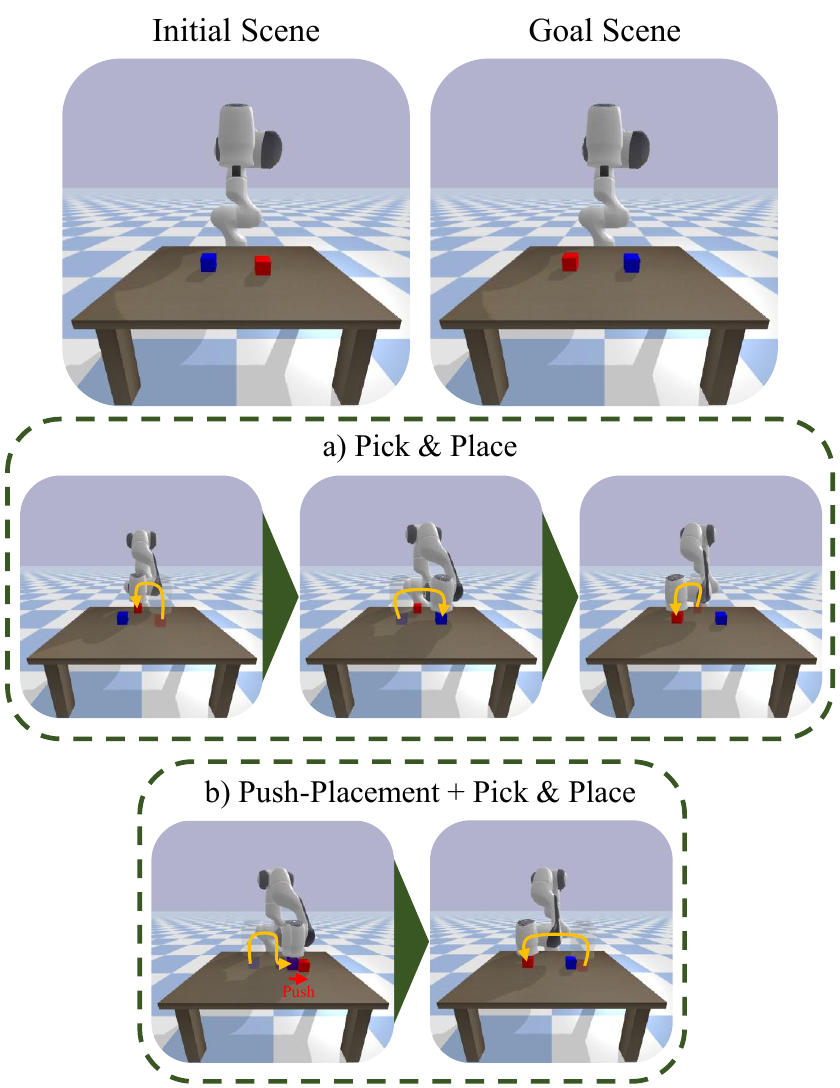}
    \caption{Two-object swap: (a) traditional pick-and-place requires buffering; (b) push-placement resolves the swap with fewer actions by pushing obstructing object(s) while placing the held object.}
    \label{fig:swap_scenario}
\end{figure}

\section{Introduction}
% Rearrangement
Object rearrangement is a fundamental task in robotics, involving the movement of objects from an initial configuration to a desired arrangement~\cite{batra2020rearrangement}. Efficient rearrangement requires planning a sequence of manipulation actions while considering geometric and collision constraints. Achieving efficient tabletop rearrangement is NP-hard due to dependencies between object poses when initial and final configurations overlap~\cite{han2018toro}. Applications range from dynamic household settings~\cite{esfahani2024relocation} to tabletop manipulation scenarios~\cite{labbe2020mcts}.

% Prehensile vs Non-prehensile
In tabletop rearrangement, manipulation strategies can generally be categorized into two paradigms: prehensile and non-prehensile manipulation. In the prehensile paradigm, objects are manipulated one by one, typically through pick-and-place actions~\cite{labbe2020mcts, barghi2025dynamicbuffers}, which rely on scene understanding to detect viable grasps~\cite{hosseini2024primitive}. However, pick-and-place may be infeasible for ungraspable objects (e.g., thin books) or when suction-based attachment fails due to deformable surfaces. Suction-based planar pushing has been used to address some of these limitations~\cite{huang2024pmmr}, but such methods still require collision-free motion planning for individual objects and can become costly in dense clutter. Conversely, the non-prehensile paradigm avoids grasping altogether: objects can be moved via contact interactions (pushing, sliding), enabling concurrent multi-object manipulations that have proven effective for large-scale rearrangement and sorting~\cite{song2020sort, ren2025nonprehensile}. Nevertheless, non-prehensile actions introduce complex object–object and robot–object dynamics that complicate precise planning and execution in the real world~\cite{ren2025nonprehensile, gao2022trlb}.

% Push Applications
Beyond rearrangement, pushing is a versatile action primitive used both as a \emph{pre-grasp} tool and for clearing clutter in tabletop and confined settings. As a pre-grasp strategy, pushing reconfigures a single object into a pose amenable to grasping (e.g., toward an edge)~\cite{elliott2016pushgrasp, zhong2025activepusher}. In densely packed tabletop scenarios, pushing acts on multiple objects to declutter or retrieve targets, ranging from full workspace clearing~\cite{zeng2018synergies} to targeted multi-object separation and retrieval~\cite{wang2024clutter, liu2022gegrasp, wu2023multitarget}. In confined retrieval and shelving, pushing clears obstructing items to enable reach or insertion: early action primitives like \emph{sweep} and \emph{push-grasp} addressed large obstacles and uncertainty~\cite{dogar2011sweep}, while later work examined simultaneous pushing during approach~\cite{dogar2013physics}, human-guided planning~\cite{papallas2020human}, topological/group pushing~\cite{vieira2024homology}, and integrated prehensile–non-prehensile pipelines~\cite{lee2025rearrangeretrieve}. Whereas these methods focus on clearing paths during approach or insertion, this work applies the idea at the \emph{placement} stage: push-placement displaces obstructing items while depositing the grasped object, thereby reducing buffering.

% Stowing
Another important application is stowing, where objects must be inserted into densely packed shelves. Here, pushes are used to create insertion space either via external mechanisms (e.g., extendable planks)~\cite{hudson2025stowing} or by using the grasped object itself to displace neighboring items during placement~\cite{chen2023stowing}. The latter realizes pushing and placement within a single continuous motion, enabling insertion into constrained spaces without prior clearance. Building on this idea, the present work generalizes such hybrid manipulation to tabletop rearrangement: push-placement is integrated into the planning process to reduce explicit buffering and improve overall efficiency.

% Buffering Challenge and Proposed Solution
A fundamental challenge in prehensile rearrangement is \emph{buffering}—temporarily moving obstructing objects when target poses are occupied. As shown in Figure~\ref{fig:swap_scenario}, resolving a two-object swap normally requires three actions: buffer one object, place the other, then retrieve the buffered object~\cite{labbe2020mcts, barghi2025dynamicbuffers}, which increases both action count and robot travel. To address this, \emph{push-placement} is introduced: a hybrid action enabling simultaneous placement and pushing during deposition. By leveraging the grasped object to clear obstructions while placing it, push-placement reduces the need for explicit buffering and lowers the number of actions for a swap from three to two. When coupled with Monte Carlo Tree Search (MCTS) planning~\cite{labbe2020mcts}, push-placement yields more efficient rearrangement sequences with reduced manipulator travel.
Although pushing has been used for stowing and decluttering, integrating prehensile and non-prehensile actions within a single action primitive—using the grasped object to push others during placement—remains largely unexplored for tabletop rearrangement. This hybrid approach, in which the held object actively manipulates neighboring items during placement, is the focus of the present work.

% physics-in-the-loop
Manipulation through pushing inherently involves uncertainty from contact dynamics and friction. As noted in~\cite{stuber2020pushsurvey}, pushed-object motion is sensitive to physical parameters that are often unknown or difficult to model. To mitigate this, physics engines are commonly used as black-box forward models. In this work, PyBullet is used in a physics-in-the-loop manner to compensate for a simplified planning model. Pushed object motion is approximated as a straight-line displacement in the push direction, and the resulting state is then verified and corrected via simulation. This iterative planning–execution loop yields robust rearrangement despite simplified push predictions while keeping planning tractable.

% Summary & Contributions
The proposed push-placement action unifies pick-and-place and pushing into a single manipulation action primitive, reducing the need for explicit buffering and improving rearrangement efficiency.
In this study, object poses are obtained from the PyBullet simulator; integrating real-time perception and deploying the method on a robot is left for future work.
The contributions of this paper are:
\begin{itemize}
    \item A hybrid manipulation framework that integrates prehensile and non-prehensile actions for efficient tabletop object rearrangement.
    \item The push-placement action primitive, which enables simultaneous placement and pushing to minimize buffering and reduce the total number of actions.
\end{itemize}

% Paper roadmap
The remainder of this paper is organized as follows. Section~\ref{sec:method} presents the push-placement action primitive for object rearrangement planning, detailing the framework and execution scheme. Section~\ref{sec:results} reports experimental results and discussion. Section~\ref{sec:conclusion} concludes the paper.

\section{Object Rearrangement Planning using Push-Placement}
\label{sec:method}

This section introduces the proposed cost-efficient framework for tabletop rearrangement and is organized as follows: (i) Formulation specifies the workspace, object footprints, goal condition, and motion-cost objective; (ii) MCTS-Based Rearrangement presents the time-bounded MCTS, state/action set, and the policy over pick-and-place and push-placement; (iii) Push-Placement and physics-in-the-loop Planning defines the push-placement action primitive and feasibility checks, and summarizes the physics-in-the-loop execution.

\subsection{Formulation}

The rearrangement problem is defined on a planar tabletop \(\mathcal{W} \subset \mathbb{R}^2\) containing a finite set of movable objects \(\mathcal{O} = \{o_1,\dots,o_N\}\). Each object \(o_i\) is an axis-aligned cuboid with center \(p_i = (x_i, y_i)\) and half-dimensions \((a_i, b_i) > 0\). Its footprint
\begin{equation}
\mathcal{F}(o_i) = [x_i-a_i,\,x_i+a_i] \times [y_i-b_i,\,y_i+b_i],
\end{equation}
must remain within the tabletop, \(\mathcal{F}(o_i) \subseteq \mathcal{W}\), and non-overlapping with all other object footprints. Given an initial arrangement \(P^0 = \{p_i^0\}\) and a target arrangement \(P^\star = \{p_i^\star\}\), object \(o_i\) is considered to have reached its goal when
\begin{equation}
\lVert p_i - p_i^\star \rVert \le \epsilon,
\end{equation}
for a small tolerance \(\epsilon > 0\). For a target object \(o_i\), an object \(o_j\) is classified as \emph{blocking} if its footprint intersects the target footprint, \(\mathcal{F}(o_j) \cap \mathcal{F}^\star(o_i) \neq \varnothing\).

Objects are assumed to have planar faces that enable stable, predictable pushing. The action set comprises pick-and-place and push-placement primitives, each required to satisfy: (i) footprints remain within \(\mathcal{W}\); (ii) unintended collisions are avoided; and (iii) intermediate and final poses remain physically stable.

Planning is geometry-based and distance-based, and does not explicitly parameterize friction. Ordinary tabletop contact conditions are assumed, under which face-to-face pushes translate blockers without uncontrolled slippage; consequently, friction does not affect the objective (end-effector travel) and enters only implicitly through execution. Distances and rotations are treated as the dominant factors.

A plan is a finite sequence of primitive actions \(\pi = (a^1,\dots,a^T)\) generating a state sequence \(P^{t+1} = f(P^t, a^t)\) that transforms \(P^0\) into a state satisfying the goal condition. The motion-cost functional
\begin{equation}
J(\pi) = \sum_{t=1}^{T} c(P^t, a^t), \qquad
c(P^t, a^t) = \lambda\, L(P^t, a^t),
\end{equation}
quantifies the end-effector travel, used in~\cite{barghi2025dynamicbuffers}, associated with the executed plan, where \(L(P^t, a^t)\) includes the approach-to-grasp, pick (grasp acquisition and lift), and transfer induced by action \( a^t \), and \(\lambda > 0\) is a scaling factor. This cost model is used as an evaluation metric for comparing rearrangement strategies. The planning procedure described in~\ref{subseb:mcts} produces feasible plans under a time budget and is not formulated as a direct minimization of \(J(\pi)\).

\begin{figure}[t]
    \centerline{\includesvg[ width = 0.5\textwidth]{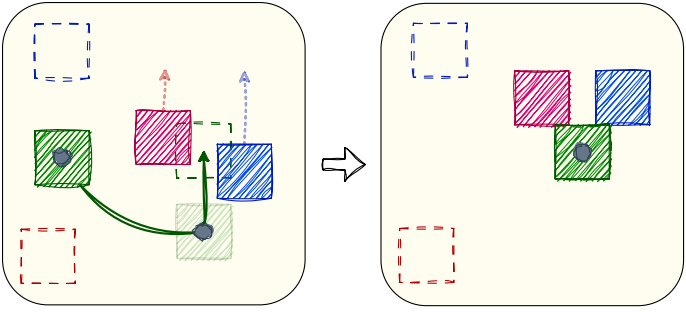}}
    \caption{Top-down example of push-placement. Colored rectangles depict current object footprints; dashed-outline rectangles (no fill) indicate target footprints. The gripper is depicted as a small filled circle on the grasped target. The target is aligned with its goal and advanced from the selected side to sweep blocking objects out of the goal region before placement.}
    \label{fig:top view}
\end{figure}

\begin{figure}[t]
    \centerline{\includesvg[ width = 0.4\textwidth]{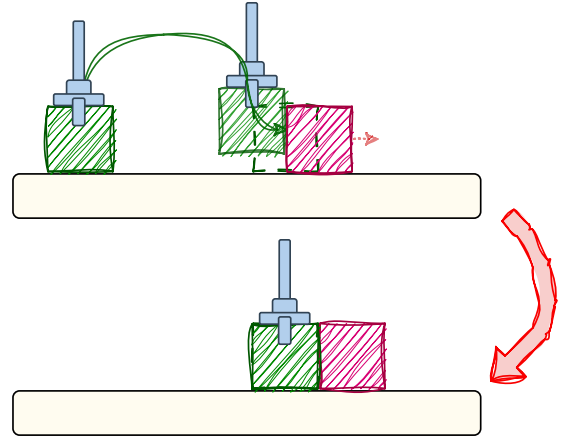}}
    \caption{Side view of push-placement. The gripper holds the target above the table and moves it along the chosen side to sweep blockers out of the goal region; the target is then placed at the goal pose once the region is clear.}
    \label{fig:side view}
\end{figure}

\subsection{MCTS-Based Rearrangement}
\label{subseb:mcts}

Rearrangement planning is cast as a stochastic tree search over scene arrangements~\cite{labbe2020mcts}. Each node represents a state \(P=\{p_i\}_{i=1}^N\), with the root at \(P^0\) and terminal nodes satisfying the goal condition \(P^\star\). Directed edges correspond to primitive actions \(a \in \mathcal{A}\) (pick-and-place or push-placement) applied via \(P' = f(P,a)\), and each edge stores the immediate motion cost \(c(P,a)\). The tree is expanded under a fixed per-instance time budget using the randomized MCTS strategy of~\cite{labbe2020mcts}.

At an expandable node \(P\), an object not yet at its target pose is sampled uniformly at random:
\begin{equation}
\label{eq:object_selection}
o \sim \mathrm{Unif}\big(\{o_i \in \mathcal{O} \mid \lVert p_i - p_i^\star \rVert > \epsilon\}\big),
\end{equation}
and an action recommender proposes a primitive for the sampled object according to the decision logic for clear and blocked targets. Expansion continues until a node satisfying \(P^\star\) is encountered or the time limit is reached. Once a goal node is found, the plan is extracted by backtracking the unique path to the root.

\begin{algorithm}[t]
\caption{Push-Side and Pre-Push Pose Selection}
\label{alg:push_side}
\begin{algorithmic}[1]
\State \textbf{Input:} Set of blockers for the target goal position $B$; current table layout
\State \textbf{Output:} Push-Side $s \in \{\mathrm{Left}, \mathrm{Right}, \mathrm{Up}, \mathrm{Down}\}$ and Pre-Push Pose $p_0$
\vspace{4pt}
\For{$s$ in $\langle \mathrm{Left}, \mathrm{Right}, \mathrm{Up}, \mathrm{Down} \rangle$}
    \State $is\_side\_appropriate \gets$ \textbf{true}
    \State $outermost\_blocker \gets$ \textbf{unset}
    \For{each blocker $b \in B$}
        \State Check that shifting $b$ along $s$ keeps it on the table
        \State Check that the rectangular corridor along $s$ from $b$ toward the far side of the goal is empty
        \If{any check fails}
            \State $is\_side\_appropriate \gets$ \textbf{false}
            \State \textbf{break}
        \EndIf
        \State Update $outermost\_blocker$ if $b$ is farther along $s$ than the current $outermost\_blocker$
    \EndFor
    \If{$is\_side\_appropriate$ and $outermost\_blocker$ is set}
        \State Form $p_0$: the position immediately behind $outermost\_blocker$ along $s$
        \State Validate $p_0$: the target footprint at $p_0$ must lie on the table and not overlap with any object
        \If{$p_0$ is valid}
            \State \Return $(s, p_0)$
        \EndIf
    \EndIf
\EndFor
\State \Return None
\end{algorithmic}
\end{algorithm}

\begin{figure*}[t]
    \centering
    \includesvg[width=0.9\textwidth]{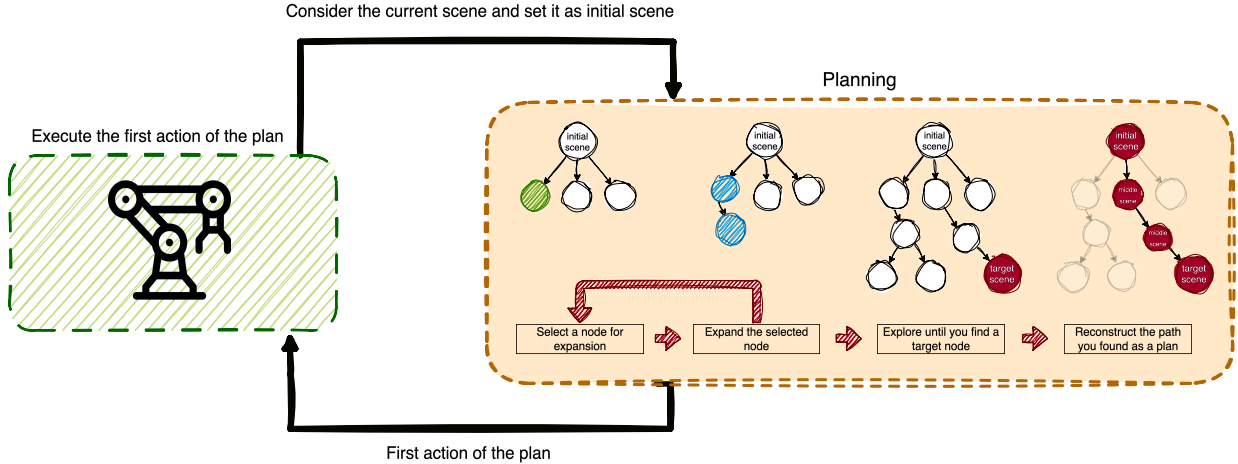}
    \caption{Physics-in-the-loop execution. At each iteration, a complete plan from the current state to the goal is computed; only its first action is executed. The updated state is then considered and the process repeats, accommodating contact-induced changes (e.g., rotations) before committing to subsequent actions.}
    \label{fig:physics-in-the-loop}
\end{figure*}

The action recommender implements the following decision logic for the chosen object \(o_i\):
\begin{enumerate}
  \item If the goal region of \(o_i\) is free (no overlap with other objects), propose a direct pick-and-place to \(p_i^\star\).
  \item If the goal region of \(o_i\) is blocked:
  \begin{enumerate}
    \item \textbf{Push-first:} test push-direction feasibility over \(s \in \{\mathrm{Left},\mathrm{Right},\mathrm{Up},\mathrm{Down}\}\) as in Algorithm~\ref{alg:push_side}. If an admissible side and a valid initial push pose exist, propose push-placement (displace blockers during the approach and place \(o_i\) at \(p_i^\star\)).
    \item \textbf{Fallback:} if no admissible push exists, sample one blocking object uniformly at random; if its goal region is free, propose pick-and-place to its target, otherwise propose pick-and-place to a buffer.
  \end{enumerate}
\end{enumerate}

The reward used during tree expansion is a state-based heuristic equal to the number of objects currently within tolerance of their targets. This heuristic biases the search toward states closer to the goal and is distinct from the motion-cost functional \(J(\pi)\), which is used only to evaluate completed plans. Because the search terminates upon reaching the first feasible plan, alternative terminal plans are not compared during a run, and motion cost is not optimized. Furthermore, the action space explored at each node is generated through randomized, time-bounded sampling of admissible actions, so the method is not complete: the existence of a feasible plan does not guarantee discovery within the bounded search time. The combination of the heuristic reward, time-bounded exploration, and first-solution termination also precludes optimality with respect to \(J(\pi)\). Stochastic action sampling may yield different plans across runs; performance is therefore reported as the mean over multiple trials. The introduction of push-placement enlarges the discrete action set and enables certain scenes to be solved with fewer primitive actions, while leaving these theoretical properties unchanged.

\subsection{Push-Placement and Physics-in-the-Loop Planning}

Push-placement executes a direct transfer of an object to its goal while sweeping the goal region to displace blocking objects. When the goal region is free, a direct pick-and-place is preferred. When the goal region is blocked and an admissible push exists, push-placement displaces blockers during the final approach, eliminating explicit pre-clearing steps and reducing both action count and travel.

The maneuver proceeds by: (i) selecting an admissible push direction from the four orthogonal sides of the goal region; (ii) positioning the target object at a pre-push pose, aligned with its goal and just outside the outermost blocker along that direction; (iii) translating the grasped target along a straight path to clear the goal region by pushing blockers into pre-validated free space; and (iv) placing the target object at the goal pose. Figure~\ref{fig:top view} shows a single-instance top-down example, and Figure~\ref{fig:side view} provides a complementary side view that highlights the gripper pose and clearance during the sweep toward the goal. A push direction is admissible if and only if, for every blocking object: (i) the required corridor and the blocker’s post-push footprint lie within the tabletop workspace; (ii) the corridor from the blocker to its post-push location is empty; (iii) the blocker remains fully on the table throughout; and (iv) no other object is contacted. 
Upon selecting $(s,p_0)$ by Algorithm~\ref{alg:push_side}, the target is grasped, moved to $p_0$, and translated along side $s$ to sweep the goal region; blockers move along $s$ into pre-validated free cells without contacting non-blockers or leaving the table, after which the target is placed at its goal pose. If no side is admissible or the pre-push pose $p_0$ is invalid, push-placement is deemed infeasible and the planner resolves the blockage via pick-and-place. 

The additional computations introduced by push-placement are dominated by the admissibility checks performed for candidate pushes. For each candidate target object the planner evaluates at most four orthogonal push directions; each direction requires only a constant-time logical evaluation of the geometric conditions (boundary, corridor emptiness, and non-interference). Hence, when the planner examines $n$ candidate targets, the total work devoted to push-direction checks grows on the order of $4n$, i.e., linear in the number of objects. In practice, if the workspace area is increased proportionally to accommodate larger $n$ (so that admissible push corridors remain possible), these per-check costs remain constant and the overall push-placement overhead scales as $O(n)$.

Execution follows a physics-in-the-loop scheme as depicted in Figure~\ref{fig:physics-in-the-loop}: after each primitive action, the updated state is considered, a complete plan to the goal is recomputed, only the first action of that plan is executed, and the cycle repeats, providing robustness to contact-induced rotations and other unmodeled effects. Contact interactions can induce rotations or small pose deviations of blockers; if the grasped target exhibits orientation drift after pickup, it is reoriented to an axis-aligned pose prior to placement. This receding-horizon procedure maintains alignment between the planner and the evolving scene.

\begin{figure}[t]
    \centering
    \includesvg[width=0.5\textwidth]{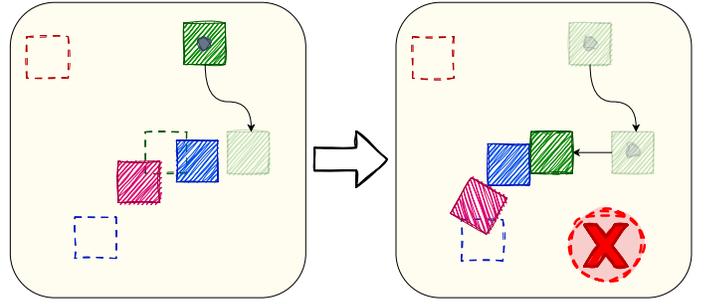}
    \caption{Unsafe chained push scenario. The grasped target would push a blocker that immediately contacts a second object, creating a contact chain that can induce large, unpredictable rotations and potentially cause a newly rotated object to occlude a previously clear goal region. Such pushes are rejected by the admissibility filter (no secondary contacts allowed). Unfilled dashed-outline rectangles indicate goal footprints.}
    \label{fig:unsafe_chain}
\end{figure}

\begin{figure}[t]
    \centering
    \includesvg[width=0.5\textwidth]{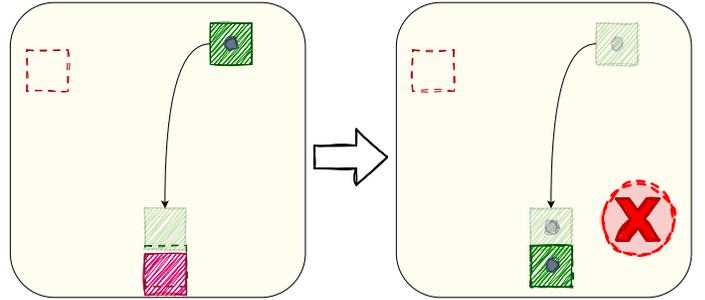}
    \caption{Unsafe edge push scenario. A blocker near the table boundary would be driven beyond the workspace if pushed along the shown direction, risking a fall. The boundary check enforces a safety margin and rejects such pushes. Unfilled dashed-outline rectangles indicate goal footprints.}
    \label{fig:unsafe_edge}
\end{figure}

The method assumes axis-aligned object footprints and relies on physics-in-the-loop execution to accommodate small pose deviations and minor rotations between actions. 
However, repeated pushes can accumulate rotation on passive objects; 
large-angle rotations violate the axis-aligned model and cannot be reliably handled within the feasibility tests. 
To mitigate this, candidate pushes are rejected that (i) would create chained contacts (illustrated in Figure~\ref{fig:unsafe_chain}, where the grasped object would push a blocker that immediately contacts another object), 
and (ii) would drive a blocker beyond a safety margin at the table boundary (Figure~\ref{fig:unsafe_edge}). These unsafe cases are filtered by the corridor-emptiness and boundary checks in Algorithm~\ref{alg:push_side}. Additionally, 
extreme rotation can cause a previously non-blocking object to newly occlude another target’s goal region (Figure~\ref{fig:unsafe_chain}); such emergent blockages are not predicted a priori but are detected after execution via physics-in-the-loop state acquisition, 
enabling replanning before further failures. Nevertheless, the approach does not explicitly reason about extreme rotations or tipping events; scenes prone to such dynamics may require reverting to pick-and-place or regrasping strategies before proceeding.

\section{Results and Discussion}
\label{sec:results}

\begin{figure}[t]
    \centering
    \includesvg[width=0.5\textwidth]{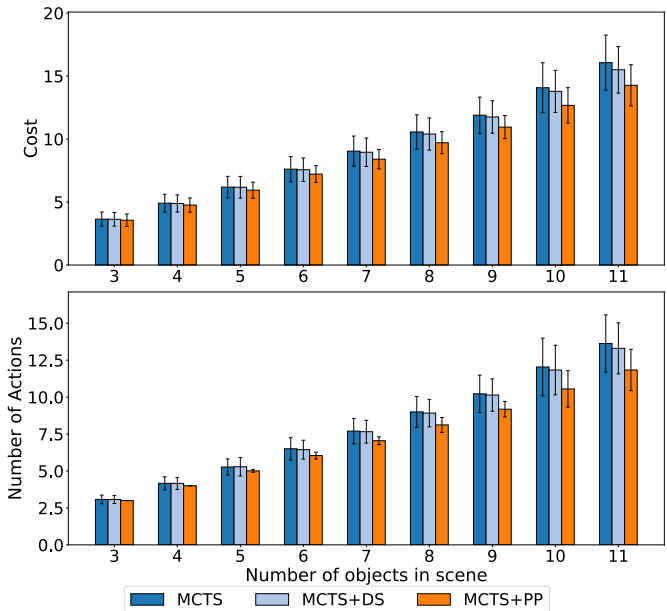}
    \caption{Mean cost and mean number of actions across 300 scenes for each object count $N$; error bars denote standard deviation across scenes.}
    \label{fig:results_bars}
\end{figure}

A time-bounded MCTS planner with a 2\,s per-instance budget was employed as the baseline configuration~\cite{labbe2020mcts}. Three action configurations were compared: (i) the standard MCTS using only pick-and-place actions~\cite{labbe2020mcts}, (ii) MCTS+DS, which integrates dynamic stacking to allow temporary object-on-object placements and improve rearrangement efficiency through reduced manipulator travel cost~\cite{barghi2025dynamicbuffers}, and (iii) MCTS+PP, which incorporates the proposed push-placement primitive to enable hybrid prehensile–non-prehensile manipulation.
For each object count $N \in \{3,\ldots,11\}$, 300 randomized tabletop scenes were generated. In each scene, objects of heterogeneous sizes were assigned random, collision-free initial and goal poses within the workspace. Each action configuration (MCTS, MCTS+DS, and MCTS+PP) was executed on all scenes with five independent runs per setup to account for variability in stochastic planning. For every scene, the number of actions and motion cost were first averaged over these five runs, providing a per-scene estimate of each metric. These per-scene averages were then aggregated across the 300 scenes for each $N$. Figure~\ref{fig:results_bars} reports the mean and standard deviation of both metrics.

Across all object counts, MCTS+PP achieves the lowest mean cost and the fewest actions, with the margin increasing as $N$ grows. The efficiency gains arise because push-placement clears the goal region during the final approach, eliminating explicit pre-clearing via buffering moves of blocking objects. This reduces the number of pick-and-place operations, shortens transfer distances, and lessens demand for buffer space. Dynamic stacking also improves over pick-and-place by providing temporary vertical buffers, but typically still requires extra transfers compared to push-placement, which consolidates clearance and placement into a single maneuver when a push corridor is admissible. When no admissible push is available, MCTS+PP reverts to standard pick-and-place, explaining convergence toward the baseline in tightly constrained scenes.

MCTS+DS incurs its own feasibility and collision checks for buffer placements and stack stability~\cite{barghi2025dynamicbuffers}; however, both MCTS+DS and MCTS+PP perform only lightweight geometric and occupancy queries per candidate action in the current implementation. Consequently, no noticeable difference in per-node computation time was observed between the two methods: both extend the action set with inexpensive feasibility tests and therefore exhibit comparable computation-time behavior under the same MCTS time limit. Overall planner runtime remains dominated by tree expansion and simulation steps; push-placement adds only a modest, linear pre-filtering cost that does not change the asymptotic search complexity.

\begin{table*}[t]
\centering
\caption{Percentage cost reduction of MCTS+PP relative to baselines (per object count).}
\label{tab:pp_gain}
\begin{tabular}{lccccccccc}
\toprule
Baseline & 3 & 4 & 5 & 6 & 7 & 8 & 9 & 10 & 11 \\
\midrule
MCTS~\cite{labbe2020mcts} & 1.65\% & 3.25\% & 3.72\% & 5.51\% & 6.56\% & 7.61\% & 7.91\% & 10.57\% & 11.12\% \\
MCTS+DS~\cite{barghi2025dynamicbuffers}       & 1.38\% & 2.65\% & 3.72\% & 5.01\% & 5.82\% & 6.81\% & 6.65\% & 8.56\% & 7.86\% \\
\bottomrule
\end{tabular}
\end{table*}

Table~\ref{tab:pp_gain} quantifies the cost advantage of push-placement. For each $N$, it lists the percentage reduction in mean cost achieved by MCTS+PP relative to MCTS and to MCTS+DS, computed as
\begin{equation}
\label{eq:reduction}
\text{Reduction} = 100 \times \frac{\bar{J}_{\text{baseline}} - \bar{J}_{\text{MCTS+PP}}}{\bar{J}_{\text{baseline}}}\,
\end{equation}
where each $\bar{J}$ is the mean cost aggregated over the 300 scenes (after per-scene averaging across five runs). The reductions grow with $N$ relative to MCTS~\cite{labbe2020mcts} (e.g., from 1.65\% at $N{=}3$ up to 11.12\% at $N{=}11$) and are consistently positive relative to MCTS+DS~\cite{barghi2025dynamicbuffers} (up to 8.56\% at $N{=}10$). These trends align with the bar plots in Figure~\ref{fig:results_bars}: push-placement delivers the largest savings by avoiding buffer relocations and concentrating clearance and placement into one action when feasible. Incorporating push-placement into planning consistently lowers both the number of actions and the cumulative motion cost, with benefits that grow in more cluttered scenes. Dynamic stacking provides a secondary improvement over pure pick-and-place, but push-placement yields the largest gains by removing the need to relocate blockers to buffers before placement.

\begin{table}[t]
    \centering
    \caption{Results in PyBullet for 100 randomized scenes (8 heterogeneous objects per scene). Success rate denotes the per-scene fraction of objects within tolerance.}
    \label{tab:pybullet_results}
    \begin{tabular}{lccc}
    \toprule
         & Robot Time & Success Rate & Num. of Actions \\
    \midrule
        MCTS & 40.79s & 97.62\% & 9.91 \\
        MCTS+PP & 38.03s & 98.50\% & 8.63 \\
    \bottomrule
    \end{tabular}
\end{table}

The proposed framework was evaluated in the PyBullet physics simulator~\cite{coumans2019pybullet} using a Franka Emika Panda model configured for 4-DoF motion. Simulation provided precise access to object states and enabled physics-in-the-loop execution without perception noise; integration of real-time object detection and pose tracking for physical deployment is left for future work. A physics-in-the-loop execution protocol was employed: at each cycle a feasible plan was produced by the time-bounded MCTS planner, only the first action of that plan was executed on the robot, the resulting state was observed, and planning resumed from the updated state. Executed primitives comprised conventional pick-and-place and push-placement actions; trials were terminated when all objects fell within a geometric tolerance of \(\epsilon=0.5\) cm of their targets or when a predefined step budget of 15 actions was exceeded. For 100 randomized scenes, each containing eight objects of heterogeneous sizes, five independent runs per scene were performed to account for stochasticity. The action count, elapsed robot time, and number of objects in target poses were recorded for each trial; these values were averaged over the runs to produce the per-scene estimates reported in Table~\ref{tab:pybullet_results}. Under these re-planning conditions, the push-placement variant (MCTS+PP) yielded lower mean action counts and shorter elapsed execution times than the pick-and-place baseline (MCTS), reflecting improved clearing of the goal region while approaching targets and thus reduced need for explicit pre-clearing relocations; a marginally higher average per-object success rate was also observed for the push-placement variant.

\section{Conclusion}
\label{sec:conclusion}
This paper presented a hybrid manipulation framework for efficient tabletop object rearrangement by integrating prehensile and non-prehensile actions through the \emph{push-placement} primitive.
Unlike traditional pick-and-place methods that require buffering when target poses are obstructed, push-placement enables the robot to push obstructing objects while placing the grasped object, 
reducing the number of required actions and the robot's travel distance. The framework was implemented within a physics-in-the-loop MCTS planner in PyBullet simulation. Experimental evaluations demonstrated that the method reduces manipulator travel cost by up to \textbf{11.12\%} compared to a baseline MCTS planner and by \textbf{8.56\%} compared to a dynamic stacking approach~\cite{barghi2025dynamicbuffers}, 
validating the efficiency of hybrid manipulation for long-horizon rearrangement tasks.

\bibliographystyle{IEEEtran}
\bibliography{bibliography}

\end{document}